\documentclass[runningheads]{llncs}
\usepackage{graphicx}

\usepackage{tikz}
\usepackage{comment}
\usepackage{amsmath,amssymb} 
\usepackage{color}
\usepackage{array}

\usepackage{gensymb}
\usepackage{subcaption}
\usepackage{bm}
\usepackage{hyperref}

\makeatletter
\newcommand{\printfnsymbol}[1]{%
	\textsuperscript{\@fnsymbol{#1}}%
}
\makeatother

\usepackage{soul}

\begin{document}
\pagestyle{headings}
\mainmatter
\def\ECCVSubNumber{100}  

\title{Beyond Weak Perspective for Monocular 3D Human Pose Estimation} 


\titlerunning{Beyond Weak Perspective for 3D Pose}
%
\author{Imry Kissos\thanks{equal contribution}\and
Lior Fritz\printfnsymbol{1} \and
Matan Goldman \and
Omer Meir \and
Eduard Oks \and
Mark~Kliger 
}
\authorrunning{I. Kissos et al.}
%
\institute{Amazon Lab126\\
\email{\{imry,liorf,matang,omermeir,oksed,markklig\}@amazon.com}}
\maketitle

\begin{abstract}
We consider the task of 3D joints location and orientation prediction from a monocular video with the skinned multi-person linear (SMPL) model. We first infer 2D joints locations with an off-the-shelf pose estimation algorithm. We use the SPIN algorithm and estimate initial predictions of body pose, shape and camera parameters from a deep regression neural network. We then adhere to the SMPLify algorithm which receives those initial parameters, and optimizes them so that inferred 3D joints from the SMPL model would fit the 2D joints locations. This algorithm involves a projection step of 3D joints to the 2D image plane. The conventional approach is to follow weak perspective assumptions which use ad-hoc focal length. Through experimentation on the 3D poses in the wild (3DPW) dataset, we show that using full perspective projection, with the correct camera center and an approximated focal length, provides favorable results.  Our algorithm has resulted in a winning entry for the 3DPW Challenge, reaching first place in joints orientation accuracy.
\end{abstract}

\begin{figure}[t]
	\includegraphics[width=\textwidth]{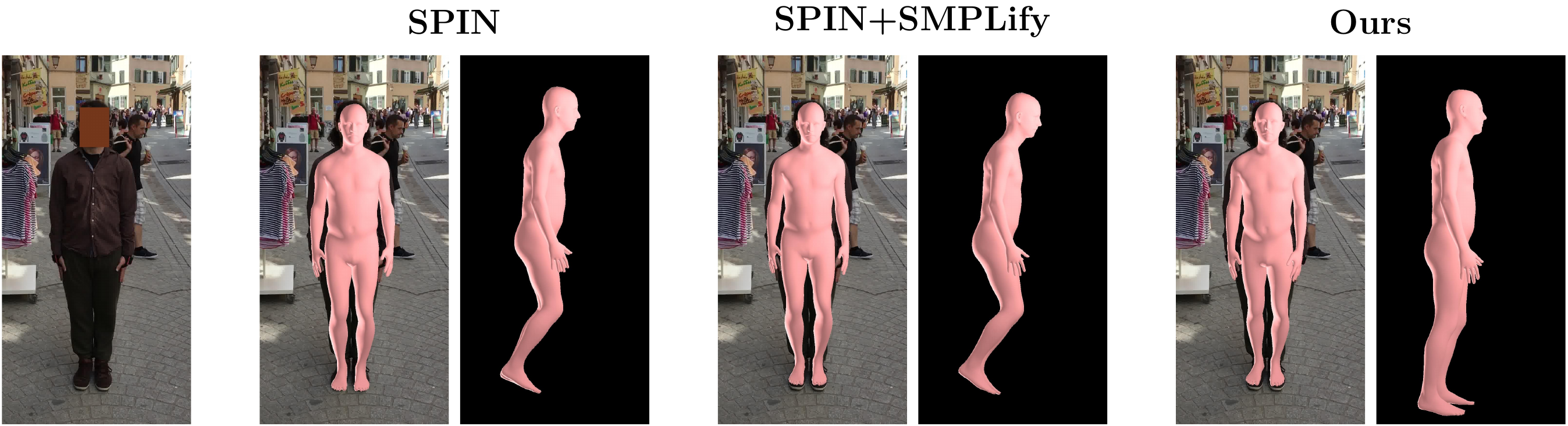}
	\centering
	\caption{The weak perspective assumption may create pose inaccuracies due to incorrect projection onto the image plane.}
	\label{fig:teaser}
\end{figure}

\section{Introduction}

Predicting 3D human joints locations from a monocular image is a challenging task. Estimating the correct depth of locations from a single image may present ambiguities which are hard to solve. Moreover, obtaining annotated data for training is a complicated and cost intensive task. In order to account for the ambiguities, prior knowledge of the human body is utilized in many models. Namely, the skinned multi-person linear (SMPL) model ~\cite{loper2015smpl}, obtained by detailed 3D scans of a large number of individuals, provides a strong prior for this task, and has been widely used in the research community~\cite{kanazawa2018end,omran2018nbf,pavlakos2018humanshape,tsungNIPS2017_7108,varol18_bodynet}.

The SMPLify algorithm~\cite{bogo2016keep}, has been shown to achieve compelling results of full 3D human mesh recovery from a single image. The algorithm fits the SMPL model to the image given locations of 2D joints on the image. The optimization is performed by minimizing an objective function that penalizes the error between projections of the estimated 3D joints onto image plane and input 2D joints. The recently proposed SPIN (SMPL oPtimization IN the loop) model~\cite{kolotouros2019learning} is a method for regressing SMPL parameters directly from an image.  SPIN has introduced a collaboration between an optimization-based SIMPLify algorithm and a deep regression network. A regressed estimate from the network initializes SMPLify with estimated SMPL parameters. The output from SMPLify serves as supervision to train the regression network. This collaboration forms a self-improving loop, so that a better network provides a better SMPLfy initialization, which leads to, again, better SMPL parameters supervision.

 In both SIMPLify optimization and SPIN deep network training, one of the main losses, so called \emph{reprojection loss}, is derived from the error between input 2D joints locations and the projections of the estimated 3D joints onto the image plane. The projection of a 3D point from the world coordinate frame into the 2D image plane requires knowledge of camera intrinsic and extrinsic parameters.  Unfortunately the camera parameters are rarely known. The authors of  SPIN have chosen the weak perspective camera model in the projection procedure. Thus, it is assumed that the camera is far from the person. This assumption is realized by using \emph{unrealistically large focal length}. Moreover, the weak perspective assumption facilitates the disregard of the actual location of the input person crop compared to the full resolution image. While this assumption may be reasonable for some data samples, mainly from the COCO dataset~\cite{lin2014microsoft}, it does not hold in some other cases, as in the 3DPW dataset~\cite{vonMarcard2018}. In fact, the weak perspective assumption may introduce noise into the reprojection loss, and in some degree, undermine its utility as a supervision signal for the estimation of 3D joints.      

To this end, we suggest to look beyond the weak perspective assumptions in the SPIN and SMPLify algorithms. We assume a full perspective projection camera model. That is, we assume that the camera is close enough to the person so that changes in depth can result in changes in the projection to the 2D image plane. We perform the projection to the original image, with respect to correct image center and approximated focal length, in contrast to the projection onto the image patch which was cropped for inference.  Additionally, we modify SMPLify camera parameters and global body orientation optimization step by leveraging all joints, rather than torso joints alone,  to compute re-projection loss which is used in fitting procedure. 

At the video level the jitter in results of frame-based 3D joints estimation can be reduced by temporal smoothing. We demonstrate that smoothing using the OneEuro filter~\cite{casiez20121}, which adaptively changes the filter cut-off frequency, produces more accurate results.

Recently, a new dataset for 3D human mesh recovery has been collected, 3D poses in the wild (3DPW) ~\cite{vonMarcard2018}. Using inertial measurement units (IMUs) and 2D joints locations, accurate full 3D human mesh, including 3D joints locations, were obtained for 60 clips in total. We experiment with this dataset, and demonstrate that the SPIN-SMPLify collaborative model with full perspective camera projection procedure achieves state-of-the-art results for joints orientations . Our algorithm was a wining entry for 3DPW Challenge reaching the first place in joints orientation accuracy.   


To summarize, the contribution of our paper is the following,
\begin{itemize}
	\item We demonstrate that using full perspective assumptions, one can achieve state-of-the-arts results for 3D joints predictions from a single image
	\item We propose to use all joints within the SMPLify camera translation and global body orientation optimization step, rather than torso joints alone.   
	\item We show that simple smoothing procedure result in better 3D joints location estimation at the video level.
\end{itemize}

\section{Related Work}

Human pose in 3D is usually represented with a set of joints positions, or with a parametric representation of a body model. Considering the task of 3D human pose inference from a single image, in the first approach, 3D joints locations are either \emph{lifted} from estimated 2D joints locations on the image ~\cite{Chen2019Unsupervised3P,martinez2017simple,moreno20173d,zhao2017simple}, 
or are directly inferred from the image  \cite{nibali20193d,sarandi2020metric}. In the second approach, several body models have been proposed, such as the SMPL model~\cite{loper2015smpl} and others~\cite{anguelov2005scape,joo2018total}. The popular SMPL model is a skinned vertex-based model that is capable of accurately representing a full mesh of the human body using a compact representation of body pose and shape parameters only.

 Recently, it has been shown that SMPL parameters can be directly inferred from an image with a deep neural network~\cite{kanazawa2018end,omran2018nbf,pavlakos2018humanshape,tsungNIPS2017_7108,varol18_bodynet}. Other methods, such as SMPLify~\cite{bogo2016keep}, consider optimizing these parameters directly given locations of 2D joints on the image. These two paradigms have been recently merged in the SPIN algorithm~\cite{kolotouros2019convolutional}, where the SMPLify algorithm is incorporated in-the-loop of the training procedure of a deep neural network, and possibly during an inference fine-tuning step. The problem of human pose extraction from videos was investigated in~\cite{arnab2019exploiting,humanMotionKanazawa19,kocabas2020vibe,alldieck2018video,mehta2020xnect}. 
 
 Our approach closely follows the SPIN method~\cite{kolotouros2019convolutional}.  To compute reprojection losses of SPIN and SMPLify, authors of ~\cite{kolotouros2019convolutional}  use unrealistically large focal length, that is inherently equivalent to weak perspective assumption.  Moreover, the reprojection losses are computed with respect to a resized cropped image of the person. In this paper we attempt to correct shortcomings of the original method related to the weak perspective assumption and projection of 3D joints onto the resized cropped image patch.  In~\cite{mono-3dhp2017}, the authors use corrective rotation to compensate for the perspective distortions from the projection to the cropped patch. However, weak perspective assumption is still used in the projection procedure.
   In~\cite{habibie2019wild}  a neural network directly predicts the parameters of the camera projection model, namely a focal length related scaling parameter and principal point. Yet, weak perspective projection is still assumed. 

\section{Method}

SPIN neural network receives as an input a cropped image $ I $ of a person, and outputs body pose and shape parameters, $ \theta $ and $ \beta $, respectively.
The network also estimates 3 camera parameters, $ (s, t_x, t_y) $. The translation parameters $ t_x$ and  $t_y$ represent the shift of the body kinematic tree root (pelvis) relatively to the origin of the cropped image coordinate frame on $ x $ and $ y $ axes respectively, while the scale parameter $ s $ is used to compute translation parameters $ t_z$  on $ z $ axis. 
The body pose parameters, $ \theta $, represent local rotation transformations of each joint compared to its parent in the kinematic tree. In total, it contains 72 parameters, 3 rotation parameters per each of the 24 joints in the kinematic tree. Additionally, 10 body shape parameters, $ \beta $, represent linear coefficients of a low-dimensional shape space, learned from a training set of thousands of registered body scans. The SMPL model uses the body pose and shape parameters to compute the locations of 6890 3D mesh vertices. Additionally, a linear joints regressor is used to infer 3D joints locations from the 3D mesh vertices.

The SMPLify algorithm requires an input of 2D joints locations. The popular choice is 25 joints defined by the OpenPose model \cite{cao2017realtime}, and we use an off-the-shelf implementation of OpenPose to obtain 2D joints locations for each image. 
The extracted 3D joints from the SMPL model are projected onto the 2D image plane using perspective projection. The joints re-projection loss together with pose and shape prior losses are minimized to estimate body pose, shape, global body orientation and camera translation parameters. The SMPLify optimization is initialized with the output of SPIN neural network.

\subsection{Projection Formulation}
\label{sec:projection_formulation}

We assume a pinhole camera model. In order to perform projection of 3D joints onto the 2D image plane, we first need to define the camera intrinsic matrix,
\begin{equation}
\label{eq:intrinsic_matrix0}
K = 
\begin{bmatrix}
f_x & s_k & o_x\\
0 & f_y & o_y\\
0 & 0 & 1
\end{bmatrix},
\end{equation}
where $ f_x, f_y $ are the camera \emph{focal length} values (in pixels), $ [o_x , o_y] $ is the camera principal point (in pixels) and $s_k$ is a skew coefficient. Usually $ f_x \equiv f_y\equiv f $, $ [o_x , o_y] \equiv  [W/2, H/2] $, where $ W $, $ H $ are the image width and height, respectively, and $s_k\equiv 0$. The intrinsic matrix can thus be rewritten as,
\begin{equation}
\label{eq:intrinsic_matrix}
K = 
\begin{bmatrix}
f & 0 & W/2\\
0 & f & H/2\\
0 & 0 & 1
\end{bmatrix}.
\end{equation}

Let us also define $ T = [t_x, t_y, t_z] \in \mathbb{R} ^ 3$ as the camera translation in the world coordinate frame. We assume that the  camera coordinate frame is aligned with the world frame, i.e., the camera extrinsic matrix is the identity matrix. Moreover, we assume that the camera does not have any radial or tangential distortion. Given a 3D point in the world coordinate system, $ p_{3D} = [x, y, z] \in \mathbb{R} ^ 3$, we compute its 2D projection onto the image plane, $ p_{2D}\in \mathbb{R} ^ 2$, with the following formulation,
\begin{equation}
\label{eq:projection}
\left[\widetilde x, \widetilde y, \widetilde z \right]^T=  K \cdot \left([x, y, z] + T\right)^T,
\end{equation}
and,
\begin{equation}
\label{eq:projection1}
p_{2D}= \left[\widetilde x/\widetilde z , \widetilde y/ \widetilde z \right].
\end{equation}

In SPIN, the person \emph{square} bounding box is first cropped, and then resized into a fixed resolution of $ 224\times224 $. The projection of 3D joints onto the image plane, and the computation of re-projection losses are performed directly onto a \emph{resized cropped image} around the person. Since the camera focal length, $ f $, is unknown, the focal length with respect to the resized cropped image patch was defined to be $f=5000 $. The  scale  parameter $ s $ is converted into the camera translation in the $z$ axis, $ t_z $, using the formula,
\begin{equation}
t_z = \frac{2 \cdot f}{Res  \cdot s},
\label{tz_1}
\end{equation}
where  $ Res \equiv 224 $ is the resized crop resolution. For most images, the resulting $ t_z $ is very large comparing to changes in the $z$ coordinate of the 3D joints. The inherent assumption here is of weak perspective projection. It is assumed that the person is very far from the camera, and that the possible changes in the $z$ coordinate of 3D joints are \emph{negligible} compared to the distance from the camera, which is not always true in practice. In fact, this assumption may lead to erroneous projection of the 3D joints onto the image plane, and as a result, a noisy reprojection loss.  Moreover, since the parameters $ (s, t_x, t_y) $ are estimated with respect to the center of the \emph{resized cropped image}, they do not truly represent the camera translation parameters of the real camera center, which is related to the center of the full resolution image.     

We propose to consider a more realistic focal length and to project 3D joints directly onto the original full resolution image plane, rather than to the resized cropped low resolution patch. In case the focal length is known or when the camera can be calibrated, the true focal length should be used. In practice, the focal length is usually unknown and the camera is unavailable for calibration. The focal length cannot be optimized as part of the global optimization process, nor can it be estimated by a neural network, since the problem is too unconstrained to optimize it together with camera translation. To overcome this problem, we suggest to use focal length approximation. 

It is known that the focal length can be calculated from the camera \emph{field of view} (FOV),
\begin{equation}
f = \frac{\sqrt{W^2+H^2}}{2 \tan{(\alpha/2)}},
\end{equation}
where $ W, H $ are the width and height of the original image and $\alpha$ is the diagonal FOV of the camera. Since the FOV of the camera is also unknown, we can simply approximate the focal length by,
\begin{equation}
\label{eq:focal_length_approx}
f \approx \sqrt{W^2+H^2}.
\end{equation}
This roughly corresponds to a camera FOV of $ 55^{\circ} $. In a $ 1920 \times1080 $ resolution camera, we get $ f\approx 2200 $. In Section~\ref{sec:focal_length} we show that the model is insensitive to the exact value of the focal length within a wide range. 

 We define $ r = b/Res $, where $b$ is the size of the detected person square bounding box in native image resolution, as the resizing factor from the  cropped image resolution to the actual input image resolution. Down-sampling the image by the factor $r$ changes the focal length of the camera (in pixels) to be $ f/r $. Since the camera scale parameter $ s $ is estimated with respect to the resized image patch, in order to convert it into the correct camera translation in the $z$ axis, $ t_z $, we use,

\begin{equation}
t_z = \frac{2 \cdot f}{r \cdot Res \cdot s}.
\label{tz_2}
\end{equation}


Finally, the estimated translation parameters $ t_x, t_y $, which are obtained from the SPIN neural network output, are related to the center of the resized cropped image and not the actual camera center at the full resolution image center. Thus, we need to perform the adequate correction.  Let $ c_x, c_y $ be the 2D coordinates of the crop center in the original full resolution image coordinates. We compute the  camera center \emph{shift} parameters using,
\begin{align}
\hat c_x = \frac{2\left(c_x - W/2\right)}{s \cdot b}&\\
\hat c_y = \frac{2\left(c_y - H/2\right)}{s \cdot b}&, \nonumber
\label{shifts}
\end{align}
where $ b $ is the bounding box size, which is always square in our case, and $ W $ and $ H $ are the original image width and height.
We use the camera center shift parameters, $ \hat c_x, \hat c_y $ to compute the camera translation parameter, $ \hat T $, with respect to the true camera location,
\begin{equation}
\hat T = \left[ t_x - \hat c_x, t_y - \hat c_y, t_z \right],
\label{hat_T}
\end{equation}
which we substitute with $ T $ in Equation \ref{eq:projection}.

The usage of the modified projection formulation is described in the next subsection.

\subsection{SMPLify Camera and Global Orientation Optimization}
\label{sec:smplify_camera_opt}
The SMPLify algorithm works in two separate optimization steps. The first, optimizes the global orientation (first 3 parameters of $ \theta $) and camera translation parameters $ T $. In the second steps, only joints orientations $ \theta $ and body shape parameters $ \beta $ are optimized. 

Since the camera translation and global orientation are important when assuming full perspective projection, we propose the following modification to the algorithm. We estimate $ (s, t_x, t_y) $ using SPIN deep regression network. We then initialize optimization of camera translation parameter $ T $ with $\hat T$ obtained from modification of $ (s, t_x, t_y) $ using equations~\ref{tz_2}-\ref{hat_T} . The fitting loss is obtained by projecting predicted 3D joints to them image plane, and comparing them with given 2D joints. In the projection procedure formulated in Equations~\ref{eq:intrinsic_matrix}-\ref{eq:projection1} we use the approximated realistic focal length as described in Equation~\ref{eq:focal_length_approx}. We also note that in~\ref{eq:intrinsic_matrix} we use the original image width and height. In SPIN, $ W $ and $ H $ are set according to the input cropped image size, so that they are both 224. 

Finally, while in~\cite{kolotouros2019learning}  this reprojection loss is computed only for the torso joints (hips and shoulders), we found it beneficial to use all joints in our case. Similarly to the joints reprojection loss used in the second step of SMPLify optimization, we weight the contribution of each joint by the squared confidence of its estimate provided by the 2D joints estimation algorithm.

\section{Experiments on 3DPW}

We experiment on the 3D poses in the wild (3DPW) dataset~\cite{vonMarcard2018}. We do not use the ground truth 2D joints or bounding box locations. Rather, we first predict 25 2D joints locations with the OpenPose algorithm~\cite{8765346,cao2017realtime}, using an off-the-shelf implementation \footnote{\href{https://github.com/CMU-Perceptual-Computing-Lab/openpose}{https://github.com/CMU-Perceptual-Computing-Lab/openpose}}. We then use these predicted joints and match them to the provided in 3DPW ground truth joints in each frame in order to track each person in the clip individually. The bounding box at each frame is obtained from the maximal and minimal values of the predicted joints coordinates.
We also use the trained SPIN model~\cite{kolotouros2019learning} to obtain an initial prediction of body shape and pose, and camera parameters. We then use SMPLify optimization, following its implementation within SPIN, with 100 iterations with a learning rate of $ 0.01 $ to optimize these parameters to match the obtained 2D joints. The SMPLify optimization is modified as was described in Section \ref{sec:smplify_camera_opt}. 

\subsection{Results}

To evaluate our model, we consider the 24 joints defined in the SMPL kinematic tree (root is used for matching).
We measure 6 metrics on the 3DPW dataset to assess our model,

\begin{itemize}
	\item \textbf{MPJPE}: mean per joint position error (in mm). Average distance from prediction to ground truth joint positions (after root matching).
	\item \textbf{MPJPE\_PA}: mean per joint position error (in mm) after Procrustes alignment (rotation, translation and scale are adjusted).
	\item \textbf{PCK}: percentage of correct joints. A joint is considered correct when it is less than 50mm away from the ground truth. The joints considered here are: shoulders, elbows, wrists, hips, knees and ankles.
	\item \textbf{AUC}: total area under the PCK-threshold curve. Calculated by computing PCKs by varying from 0 to 200 mm the threshold at which a predicted joint is considered correct.
	\item \textbf{MPJAE}: measures the angle in degrees between the predicted part orientation and the ground truth orientation. The orientation difference is measured as the geodesic distance in SO(3). The 9 parts considered are: left/right upper arm, left/right lower arm, left/right upper leg, left/right lower leg and root.
	\item \textbf{MPJAE\_PA}: measures the angle in degrees between the predicted part orientation and the ground truth orientation after rotating all predicted orientations by the rotation matrix obtained from the Procrustes alignment step. 
\end{itemize}

We compare ourselves to the results of SPIN~\cite{kolotouros2019learning} and SPIN followed by fine-tuning with its original SMPLify implementation. Table~\ref{table:results_ours} summarizes the consistent improvements over both methods across all metrics. The modifications to the camera optimization listed in Section~\ref{sec:smplify_camera_opt} lead to improvements in MPJPE and MPJAE from 84.174 to 83.154 and from 20.437 to 19.697, respectively.

\setlength{\tabcolsep}{4pt}
\begin{table}
	\begin{center}
		\caption{Results on 3DPW  with unknown ground truth person crop. SPIN refers to results obtained from the pretrained network from~\cite{kolotouros2019learning}, and SPIN+SMPLify refers to fine-tuning the results with the original SMPLify implementation. Our results depict SPIN fine-tuned with our SMPLify implementation. \emph{**Our MPJAE is a current SOTA on the 3DPW dataset.} }
		\label{table:results_ours}
		\begin{tabular}{llll}
			\hline\noalign{\smallskip}
			Metric &  SPIN~\cite{kolotouros2019learning} &SPIN+SMPLify~\cite{kolotouros2019learning} & Ours \\
			\noalign{\smallskip}
			\hline
			\noalign{\smallskip}
			MPJPE $ (\downarrow) $ & 99.402 & 95.839 & $ \bm{83.154} $\\
			{MPJPE\_PA} $ (\downarrow) $& 68.131 & 66.390 & $ \bm{59.703} $\\
			PCK $ (\uparrow) $& 30.846& 33.264 & $ \bm{42.419} $\\
			AUC $ (\uparrow) $& 0.534 & 0.550 & $ \bm{0.623} $\\
			MPJAE $ (\downarrow) $& 24.380 & 23.900 & $ \bm{19.697^{**}} $\\
			{MPJAE\_PA} $ (\downarrow) $& 21.198 & 24.410 & $ \bm{19.149} $\\
			\hline
		\end{tabular}
	\end{center}
\end{table}
\setlength{\tabcolsep}{1.4pt}


\subsection{Effect of Focal Length}
\label{sec:focal_length}

We experiment with a set of a few focal length values. We note that the camera focal length used in the computation of the camera translation parameters in Equation \ref{tz_2} is normalized by the resizing factor, $ r $.  In the weak perspective implementation of SPIN~\cite{kolotouros2019learning}, the  focal length of the cropped and \emph{resized} image was assumed to be $ 5000 $,  which is equivalent to the focal length $5000 \cdot r$ on the full resolution image. As $ r $ is usually around 3, the focal length which was effectively assumed in the original SPIN implementation is about 15000.   

\begin{figure}
	
	\begin{subfigure}[t]{0.5\textwidth}
		\includegraphics[width=\textwidth]{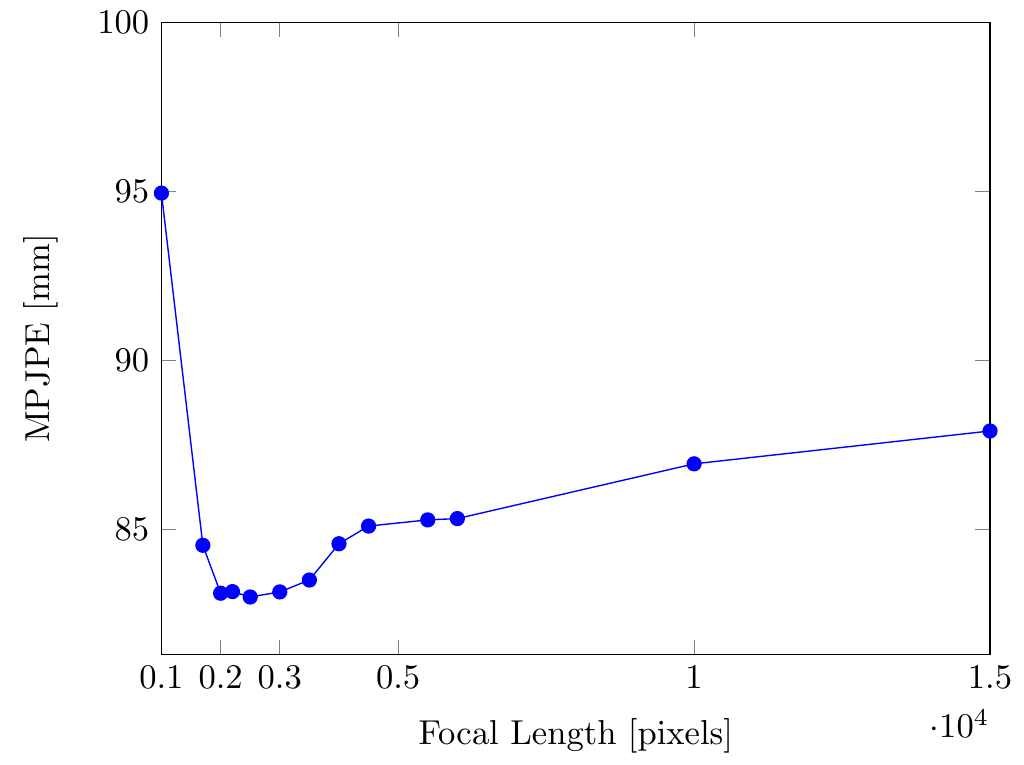}
		\centering
		\caption{}
		\label{fig:focal_length_MPJPE}
	\end{subfigure}
	~
	\begin{subfigure}[t]{0.5\textwidth}
		\includegraphics[width=\textwidth]{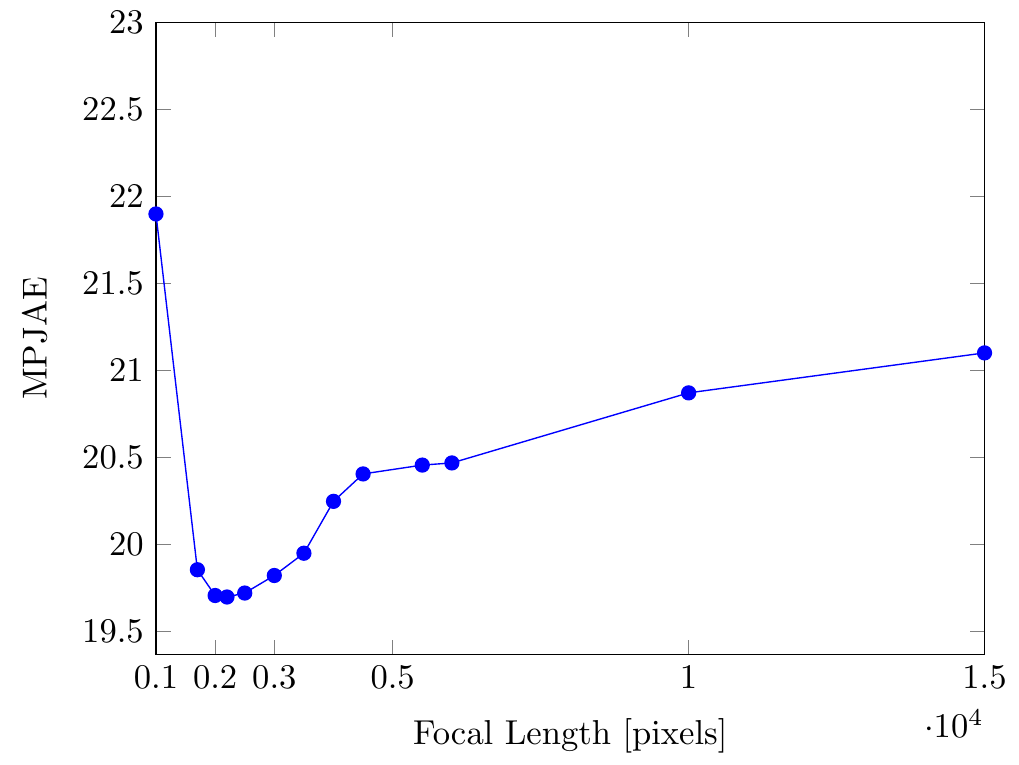}
		\centering
		\caption{}
		\label{fig:focal_length_MPJAE}
	\end{subfigure}
	\caption{In~(\subref{fig:focal_length_MPJPE}) we show the effect of the focal length on the joints locations metrics, measured in mm. In~(\subref{fig:focal_length_MPJAE}) we show the effect on the orientations metric, MPJAE, which is the geodesic distance measured between predicted and ground truth orientations, in SO(3).}
	\label{fig:focal_length}
\end{figure}

We present the effect of the focal length in pixels units on the joints distance metrics (MPJPE) and joints orientations distance (MPJAE). As can be observed in Figure~\ref{fig:focal_length}, for a large range of focal length values, the results are maintained. However, as the focal length becomes much smaller or larger, the results are deteriorated substantially. We deduce that a close approximation of the focal length is crucial, however the exact value does not have to be known.

\subsection{Effect of Camera Center}

In the original SPIN-SMPLify implementation the camera is assumed to be centered at the bounding box center. As detailed in Section~\ref{sec:projection_formulation}, we alter the optimization process so that the camera center is at the actual full resolution image center. In Table~\ref{table:camera_center} we present the effect of altering the camera center on the measured metrics. We observe consistent improvements across all metrics.

\setlength{\tabcolsep}{4pt}
\begin{table}
	\begin{center}
		\caption{Effect of using different camera center definitions on metrics}
		\label{table:camera_center}
		\begin{tabular}{lll}
			\hline\noalign{\smallskip}
			Metric & Camera Center at Bounding Box & Camera Center at Image Center\\
			\noalign{\smallskip}
			\hline
			\noalign{\smallskip}
			MPJPE $ (\downarrow) $ & 86.103 & $ \bm{83.154} $\\
			{MPJPE\_PA} $ (\downarrow) $& 59.910 & $ \bm{59.703} $\\
			PCK $ (\uparrow) $& 39.208 & $ \bm{42.419} $\\
			AUC $ (\uparrow) $& 0.607 & $ \bm{0.623} $\\
			MPJAE $ (\downarrow) $& 20.586 & $ \bm{19.697} $\\
			{MPJAE\_PA} $ (\downarrow) $& 19.181 & $ \bm{19.149} $\\
			\hline
		\end{tabular}
	\end{center}
\end{table}
\setlength{\tabcolsep}{1.4pt}

\subsection{SMPLify Number of Iterations}

\begin{figure}[h!]
	
	\begin{subfigure}[t]{0.5\textwidth}
		\includegraphics[width=\textwidth]{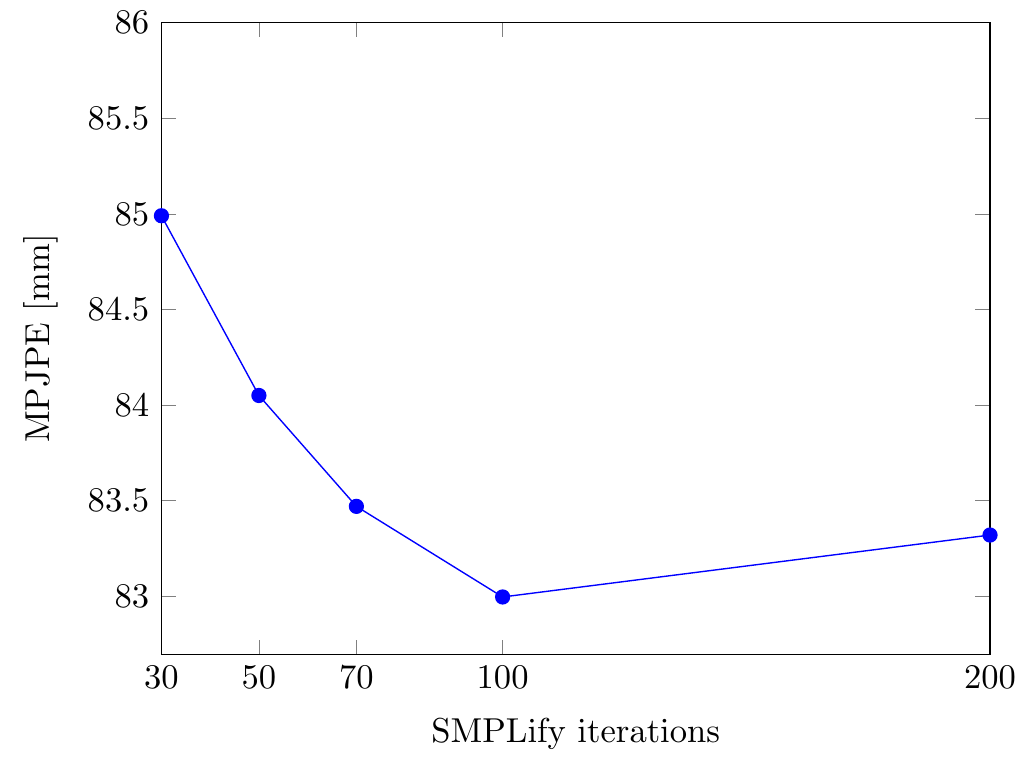}
		\centering
		\caption{}
		\label{fig:smplify_iters_MPJPE}
	\end{subfigure}
	~
	\begin{subfigure}[t]{0.5\textwidth}
		\includegraphics[width=\textwidth]{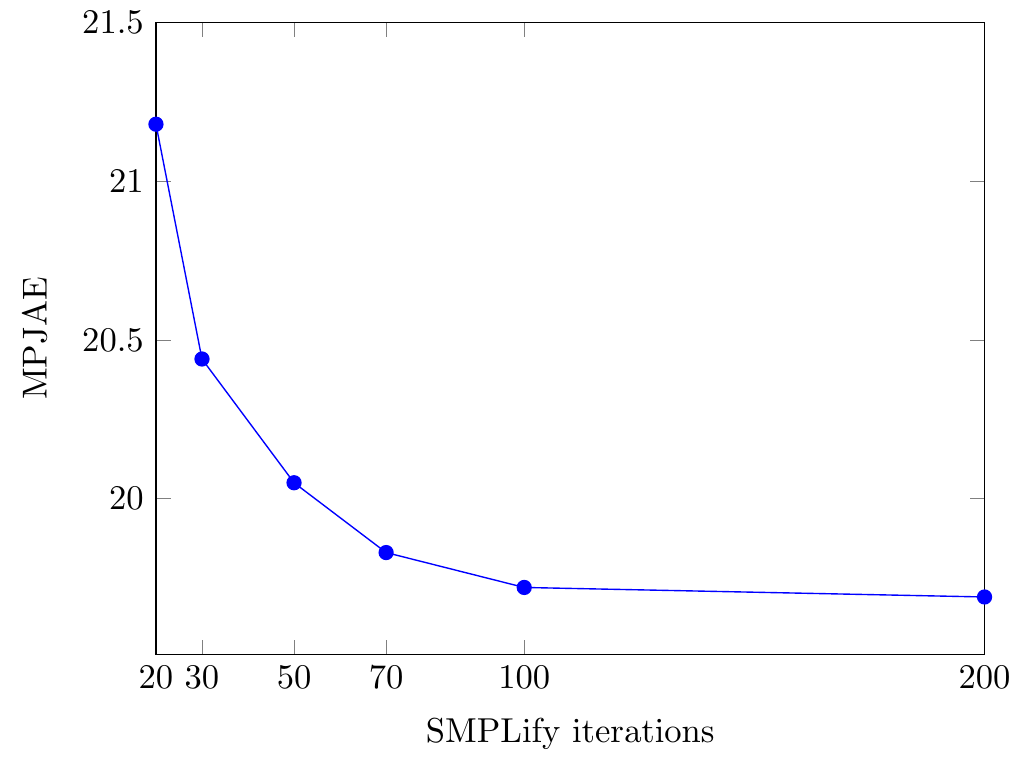}
		\centering
		\caption{}
		\label{fig:smplify_iters_MPJAE}
	\end{subfigure}
	\caption{In~(\subref{fig:smplify_iters_MPJPE}) we show the effect of the number of SMPLify optimization iterations on the joints locations metrics, measured in mm. In~(\subref{fig:smplify_iters_MPJAE}) we show the effect on the orientations metric, MPJAE, which is the geodesic distance measured between predicted and ground truth orientations, in SO(3).}
	\label{fig:smplify_iters}
\end{figure}

We experiment with different number of iterations in the SMPLify optimization process. Figure~\ref{fig:smplify_iters} shows the number of SMPLify iterations and the effect on the MPJPE and MPJAE metrics. We observe that as we increase the number of iterations, the metrics improve consistently, while at 200 iterations there is slight degradation of MPJPE.

\subsection{Smoothing}

At the video clip level, we observe considerable jitter in the predicted results. As a final post processing steps, we perform temporal smoothing on the 3D joints locations and orientations with the OneEuro filter~\cite{casiez20121}. The advantage of this filter, is that using adaptive cut-off frequency, it maintains small lag in rapid movements, while reducing noise and jitter in slow movements. We summarize the effect of smoothing in Table~\ref{table:smoothing}. While the difference is marginal, it is consistent across all metrics. We note that except of joints and orientations temporal smoothing, we do not enforce any other kinematic, skeletal or shape temporal constraints at the video level. Incorporating additional temporal constraints will be a subject for future work.

\vspace{-20pt}
\setlength{\tabcolsep}{4pt}
\begin{table}
	\begin{center}
		\caption{Effect of using smoothing with OneEuro filter on metrics}
		\label{table:smoothing}
		\begin{tabular}{lll}
			\hline\noalign{\smallskip}
			Metric & Without Smoothing & With Smoothing\\
			\noalign{\smallskip}
			\hline
			\noalign{\smallskip}
			MPJPE $ (\downarrow) $ & 84.986 & $ \bm{83.154} $\\
			{MPJPE\_PA} $ (\downarrow) $& 60.677 & $ \bm{59.703} $\\
			PCK $ (\uparrow) $& 42.182 & $ \bm{42.419} $\\
			AUC $ (\uparrow) $& 0.619 & $ \bm{0.623} $\\
			MPJAE $ (\downarrow) $& 20.063 & $ \bm{19.697} $\\
			{MPJAE\_PA} $ (\downarrow) $& 19.435 & $ \bm{19.149} $\\
			\hline
		\end{tabular}
	\end{center}
\end{table}
\setlength{\tabcolsep}{1.4pt}

\subsection{Examples from 3DPW}

\begin{figure}
	\includegraphics[width=\textwidth]{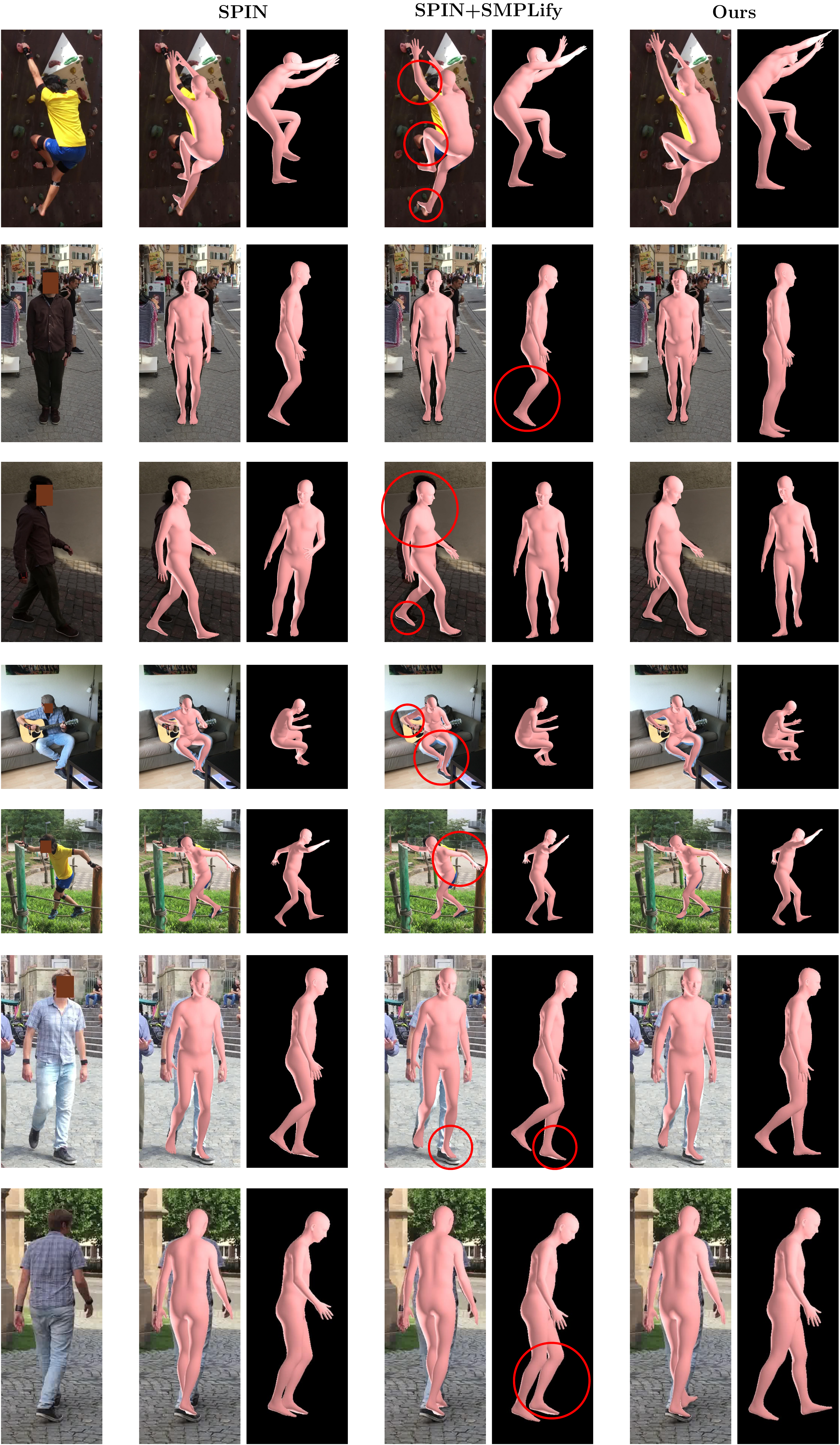}
	\centering
	\caption{Examples of some mesh recoveries from frames in the 3DPW dataset. We present results of SPIN, SPIN with the original SMPLify fine-tuning (SPIN+SMPLify) and our results. Points of interest are marked on the images.}
	\label{fig:examples}
\end{figure}

In Figure~\ref{fig:examples} we show rendered human mesh recovered by our model, and with SPIN (with and without SMPLify fine-tuning) on several extracted frames from the 3DPW dataset. Rendering is done with the same focal length that is used for optimization.

\section{Conclusions}

We have presented the effect of using full perspective projection instead of weak perspective projection within the SMPLify optimization step of SPIN method. Experiments on the 3DPW dataset show that this method provides considerable improvements across all metrics evaluated. While the full perspective projection requires knowledge of the camera focal length, it can be roughly approximated from the image resolution, and we have shown that the model is insensitive to the exact value of the focal length, and that the results are maintained within a wide range of values. Moreover, we have also shown the importance of using the correct camera center, temporal smoothing and provided some modifications to the SMPLify loss function to accommodate the new optimization scheme. As future work, one can consider removing the weak perspective assumptions from the network training, so that the results can be maintained within a single forward pass and incorporate additional temporal kinematic, skeletal or shape constraints.

\section*{Acknowledgements}

We thank the anonymous reviewers whose comments and suggestions helped improve and clarify this manuscript. We also thank our colleagues from Amazon Lab 126,  Dr. Ilia Vitsnudel, Eli Alshan, Ido Yerushalmi, Liza Potikha, Dr. Ianir Ideses and Dr. Javier Romero from Amazon Body Labs, for multiple useful discussions.     

%
%
\bibliographystyle{splncs04}
\bibliography{egbib}
\end{document}